\documentclass[conference]{IEEEtran}
\IEEEoverridecommandlockouts

\usepackage{cite}
\usepackage{amsmath,amssymb,amsfonts}
\usepackage{algorithmic}
\usepackage{graphicx}
\usepackage{textcomp}
\usepackage[dvipsnames]{xcolor}
\usepackage{url}
\usepackage{soul}
\usepackage{xspace}
\usepackage{ulem}
\usepackage{multirow,tabularx}
\usepackage{subfigure}
\usepackage{arydshln}
\usepackage{booktabs}

\usepackage[dvipsnames]{xcolor}

\title{Few-shot Domain-Adaptative Visually-fused Event Detection from Text}

\author{\IEEEauthorblockN
{Farhad Moghimifar, Fatemeh Shiri, \\
 Reza Haffari, Yuan-Fang Li} 
\IEEEauthorblockA{\textit{Department of Data Science and AI} \\
\textit{Faculty of Information Technology} \\
\textit{Monash University}, Melbourne, Australia \\
firstname.lastname@monash.edu\vspace{-20pt}\vspace{-5pt}}
\and
\IEEEauthorblockN{Van Nguyen}
 \IEEEauthorblockA{\textit{Information Sciences Division} \\
\textit{Defence Science and Technology Group}\\
Adelaide, Australia \\
Van.Nguyen5@defence.gov.au}}

\newcommand{\model}{VF-Event\xspace}
\newcommand{\im}{Visual Imaginator\xspace}

\begin{document}
\maketitle

\begin{abstract}

Incorporating auxiliary modalities such as images into event detection models has attracted increasing interest over the last few years. The complexity of natural language in describing situations has motivated researchers to leverage the related visual context to improve event detection performance. 
However, current approaches in this area suffer from data scarcity, where a large amount of labelled text-image pairs are required for model training. Furthermore, limited access to the visual context at inference time negatively impacts the performance of such models, which makes them practically ineffective in real-world scenarios.
In this paper, we present a novel domain-adaptive visually-fused event detection approach that can be trained on a few labelled image-text paired data points. Specifically, we introduce a visual imaginator method that synthesises images from text in the absence of visual context. Moreover, the imaginator can be customised to a specific domain. In doing so, our model can leverage the capabilities of pre-trained vision-language models and can be trained in a few-shot setting. This also allows for effective inference where only single-modality data (i.e.\ text) is available. The experimental evaluation on the benchmark M2E2 dataset shows that our model outperforms existing state-of-the-art models, by up to 11 points. 

\end{abstract}


\section{Introduction}

Event detection is a sub-task of event extraction in Natural Language Processing~(NLP) that aims at extracting information about events from the text. The main challenge of this task is rooted in the complexity of the natural language. Over the years, a wide variety of approaches, from rule-based methods~\cite{valenzuela2015domain,cunningham2011text,jenhani2016hybrid,freitag2016feature} to end-to-end neural network-based models~\cite{hsu2022degree,zhang2022transfer,xu2022extracting,yao2022schema,shiri2022paraphrasing} have been proposed to address this task. While these techniques have shown promising results on the existing standard benchmarks, they are heavily reliant on a single modality, i.e.\ text~(Figure \ref{fig:intro-a}), and fail to account for the real-world grounding in other related modalities. 

\begin{figure}[t]
    \centering
    \resizebox{.4\textwidth}{!}{
    \includegraphics{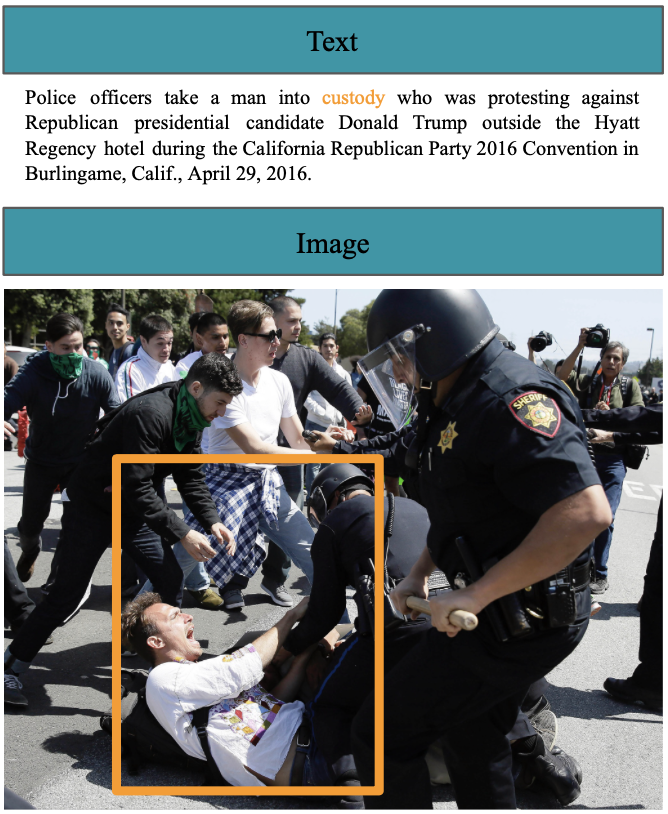}}
    \caption{An example of Multimedia VOA news from the M2E2 dataset~\cite{li2020cross}. The textual input in this article reports an event of the type Arrest:Jail, which is complemented by the visual context. The trigger word for this event is "custody", and the bounding box in the accompanying image highlights the man being taken into custody.}
    \label{fig:open}
\end{figure}

\begin{figure*}[ht]
\centering
\subfigure[]{\label{fig:intro-a}\scalebox{1}[1]{\includegraphics[width=0.3\linewidth]{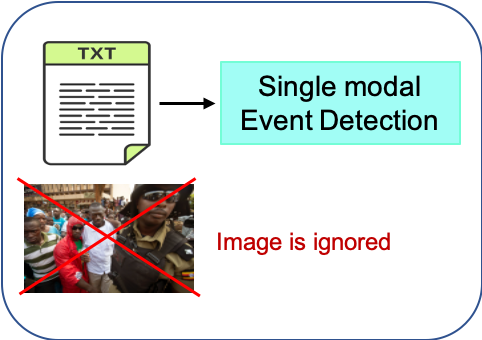}}}
\centering
\subfigure[]{\label{fig:intro-b}\scalebox{1}[1]{\includegraphics[width=0.3\linewidth]{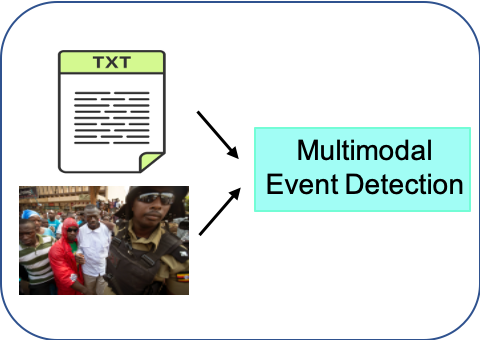}}}
\subfigure[]{\label{fig:intro-c}\scalebox{1}[1]{\includegraphics[width=0.3\linewidth]{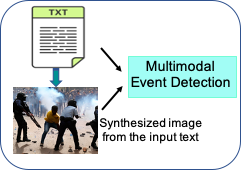}}}
\caption{The comparison among (a) conventional text-only event detection, (b) multimodal supervised event detection, and (c) our proposed end-to-end few-shot visually fused event detection model.}
\label{fig:intro}
\end{figure*}

Utilising other modalities such as images to introduce salient information that complements what is contained in text has shown to be effective in other areas such as Machine Translation~\cite{calixto2017doubly,ive2019distilling,zhu2019dm}. However, developing \emph{multimodal} 
approaches for event detection is still an under-explored area.
Consider Figure \ref{fig:open} as an example from the M2E2 dataset~\cite{li2020cross}. 
The term ``custody'' in the input text could indicate the occurrence of an \emph{Arrest:Jail} event type, as it implies the action of arresting someone. This interpretation is complemented by the visual features embodied in the image. Generally, incorporating the underlying visual representation of a situation can improve the data efficiency of event detection methods and help to uncover linguistically indescribable or missed events. 

One of the main challenges in developing such multimodal event detection models is the lack of large amounts of labelled data, which is especially critical for fine-tuning techniques based on large language models~\cite{li2020cross}~(Figure \ref{fig:intro-b}).
This issue hinders the real-world practicality where domain-specific multi-modal data (i.e.\ paired text and image) may be scarce. 
Additionally, accessing visual context in real-world scenarios might be limited, which can also affect the performance of a multimodal event detection model at inference time.

To overcome these limitations, we propose a few-shot visually-fused model for event detection that can effectively handle data scarcity during both training and inference. During training, we address the problem of having access to only a limited number of labeled image-text pairs with a few-shot transfer learning model. We initialize the parameters of our model by leveraging a pre-trained large vision-language model and update them to accommodate the distribution of the target data, thus addressing the targeted task more effectively.

In addition, we introduce a customised \emph{\im} model that can learn to synthesise visual representations of the textual input using only a few examples during training. This approach reduces the domain shift between the data that the pre-trained model was developed on and our target dataset, enabling our proposed \im to generate customised domain-specific images that complement the textual input. This approach helps us address the challenge of having access to single-modality data, i.e., text. We contemplate that our model can leverage synthetically generated images in lieu of missing actual images during inference to perform multimodal event detection. As a result, our proposed approach for event detection can achieve accuracy comparable to that of multimodal models where both text and images are fully available. 
 


In summary, our contributions are as follows:
\begin{itemize}
\item We formulate the multimodal event detection task as a few-shot learning problem and propose a novel visually-fused event detection framework, alleviating the issue of data scarcity. 
\item Our proposed customised \im module can synthesise domain-specific images using textual input to address the challenge of missing text-image pairs at inference time.
\item Experiments on a standard benchmark dataset for event detection confirm the superiority of our model over state-of-the-art models in low-resource and few-shot scenarios.
\end{itemize}

\section{Background}
Our approach is related to three lines of research.
\subsection{Multimodal Event Detection}
Most event extraction methods extract event information from single-modal data, mainly text, and ignore the contribution of visual media \cite{lu2021text2event,liu2022dynamic,shiritcg,wu2022kc}. However, in real-world scenarios, data is often multimodal. For instance, news articles, policy/regulation documents, intelligence reports and social media posts typically contain text and accompanying images. A number of event extraction methods have brought forth encouraging results by retrieving additional related visual knowledge \cite{tong2020image,zhang2017improving}. 

The problem of event extraction from multimodal data has not been investigated until recently \cite{li2020cross}, and it remains an under-explored area. 
\cite{li2020cross} introduced a multimedia structured common space construction method to take advantage of the existing image-caption pairs and single-modal annotated data for weakly supervised training.
\cite{zhang2017improving} showed that including the visual context results in better extraction performance. 
\cite{tong2020image} proposed a dual recurrent multimodal model to conduct deep interactions between images and sentences for modality features aggregation.
\cite{li2022clip} proposed an approach based on pre-trained vision-language models~\cite{radford2021learning} for addressing multimodal event extraction. However, these approaches require a complete set of text-image pairs for training. Furthermore, these models suffer from the data scarcity issue, which is especially critical in low-resource situations. Unlike these approaches, we propose a model capable of synthesising the visual context to alleviate the problem of \emph{low-resource} data.
To tackle ambiguities, one may draw on visual clues to provide clarity. We propose an automatic event detection approach, which leverages visual information.

\subsection{Vision-language Pretraining models}
Recent years, vision-language pretraining models have been widely studied~\cite{chen2020uniter,zhou2020unified,tan2019lxmert,huang2021seeing,jia2021scaling,li2020oscar,zhuo2023robustness}. It has been shown that visual, structural information such as scene graphs \cite{yu2021ernie} is useful to pretraining these models. However, event structural knowledge is not well captured in pretraning models, resulting in deficiencies in tasks related to verb comprehension \cite{hendricks2021probing}. 
Following the vision-language pretrained
model CLIP \cite{radford2021learning},
\cite{li2022clip} encoded structural event semantics and knowledge to enhance vision-language pretraining.

\subsection{Text-to-Image Generation}
Generating images from text has been extensively studied \cite{ramesh2021zero,esser2021taming,reed2016generative}. Representative works use GANs \cite{reed2016generative,xu2018attngan,zhang2021cross,yu2019can,shiri2019identity} to synthesize photo-realistic scenes with high semantic fidelity to their conditioned text descriptions. DALL-E \cite{ramesh2021zero} proposes an autoregressive Transformer with discrete VAEs \cite{van2017neural} to create images from text for a wide range of concepts expressible in natural language. 
\cite{ho2020denoising} proposed to generate high-quality images using diffusion models by establishing connections between diffusion models and variational inference.
While our approach is inspired by these works, the goal of the present work is to synthesise discrete visual representations for improving event detection instead of generating high-quality photo-realistic images.

\section{Methodology}\label{sec:approach}
\subsection{Architecture}
\begin{figure*}
    \centering
    \begin{tabular}{c}
    \includegraphics[width=1\linewidth]{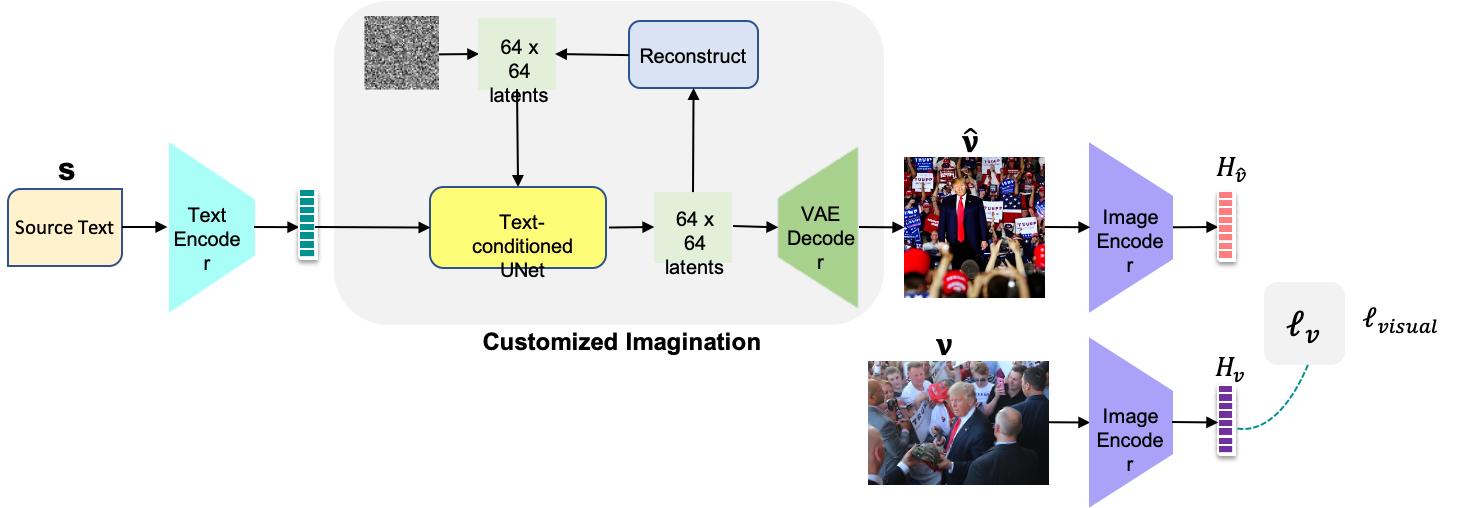}\\
    \hdashline
    \includegraphics[width=0.97\linewidth]{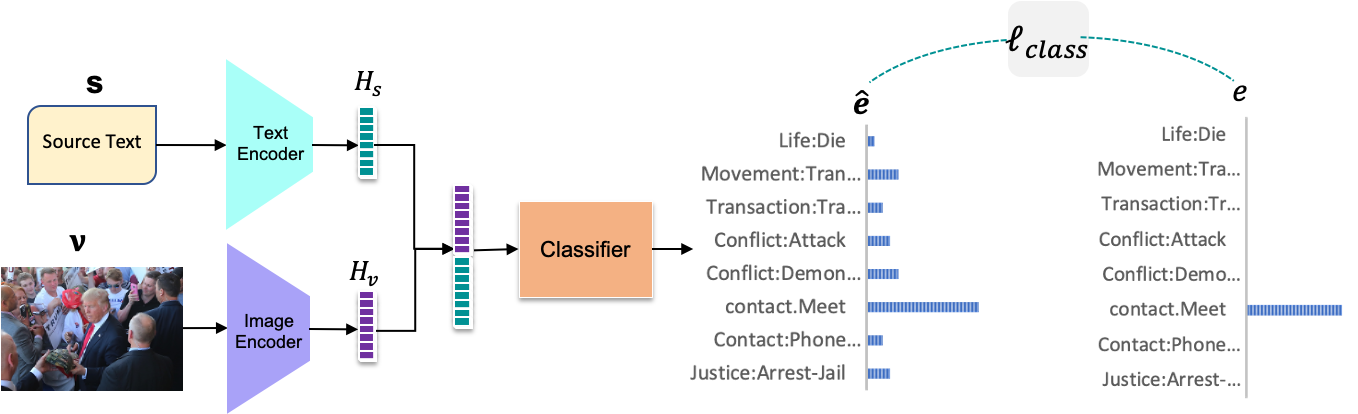}\\
    \hdashline
    \includegraphics[width=1\linewidth]{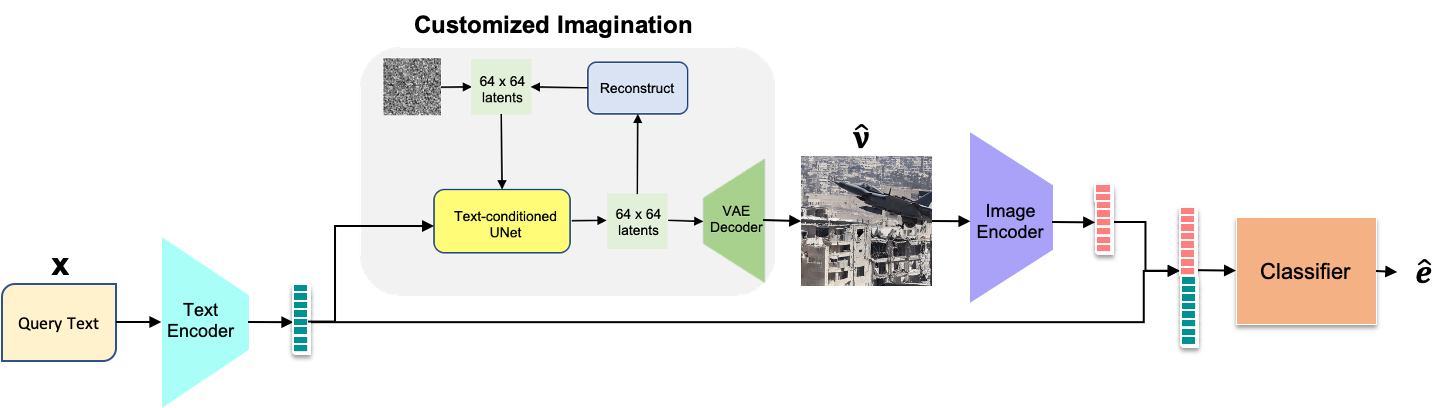}  \\
    \end{tabular}
    \caption{An illustration of our \model framework. \textbf{Top:} Customizing the imagination module with samples from support set. \textbf{Middle:} Fine-tuning of the vision-language module with the support set. \textbf{Bottom:} Inference process of \model in the absence of visual inputs.}
    \label{fig:pipeline}
\end{figure*}

Figure \ref{fig:pipeline} illustrates the architecture of our Visually-Fused Event Detection (\model) framework, which consists of two key components: (1) the visual imagination module and (2) the pre-trained vision-language module. The customized imagination module can address the challenges of multimodal event detection without requiring images at inference time.

\subsection{Task Formulation}
Our model learns to extract the event type of a query instance $s$ given a support set $\mathbb{S}$ and a set of pre-defined event types $\mathcal{E}$, which appears in the support set $\mathbb{S}$. We employ a $N$+1-way, $K$-shot setting, in which there are $N$ clusters representing $N$ event types and an additional cluster representing the \textit{none} event type, and each cluster contains $K$ data points. In addition, there are corresponding visual contexts $\mathbb{V}$ that are paired with instances in $\mathbb{S}$. The support set is denoted as follows:

\begin{align}
    \mathbb{S}=\{&(s_1^1,v_1^1,e_1),\dots,(s_1^K, v_1^K,e_1),\\
    &\nonumber \dots\\
    \nonumber
    &(s_N^1,v_N^1,e_N),\dots,(s_N^K, v_N^K,e_N),\\
    \nonumber
    &(s_{N+1}^1,v_{N+1}^1,e_{none}),\dots,(s_{N+1}^K, v_{N+1}^K,e_{none})\}
    \label{eq:support_set}
\end{align}

where:
\begin{itemize}
    \item each $s_i^j = (w_1, w_2,\dots)$ represents the $j$-th instance of the support set of event type $e_i$, and $w_m$ is a token from the vocabulary $\mathcal{W}$.
    \item $\mathbb{V}=\{v_1^1,\dots,v_1^K,\dots,v_N^K\}$ are visual contexts. $v_i^j\in \mathbb{V}$ is the  image paired with the $j$-th instance of even type $e_i$ in the support set. 
    \item $\mathcal{E}=\{e_1,\dots,e_N,e_{none}\}$ s the set of labels, representing the pool of all possible event types, and $e_{none}$ is a special label for non-event.
\end{itemize}

\subsection{Event Detection}
We formulate the task of event detection as a multimodal classification task, where our proposed model outputs the event type $e$ with the highest probability among all possible event types $\mathcal{E}$.

The encoder $\textbf{ENC}(\cdot)$ consists of a text encoder ($\textbf{ENC}_{L}$) and a visual encoder ($\textbf{ENC}_{V}$). $\textbf{ENC}_{L}$ computes the hidden representations (${H}_s$) of the input instance $s$ and a $\textbf{ENC}_{V}$ compute the visual hidden representations (${H}_v$) of the input image $v$.

Hence, the hidden representation~(${H}$) of the input $(s,v)$ is formulated as:
\begin{equation}
    {H} = \textbf{ENC}(s,v) = [\textbf{ENC}_L(s);\textbf{ENC}_V(v)]
\end{equation}
The $\textbf{FNN}(\cdot)$ is a feedforward neural network with a soft-max layer which computes the probability distribution over the possible event types:
\begin{equation}
    P_{\theta_{class}}~(e|s,v) = \textbf{FNN}_{\theta_{FNN}}(w_{class}.\textbf{ENC}_{\theta_{ENC}}(s,v)+b_{class}),
\end{equation}
where $\textbf{ENC}_{\theta_{\textbf{ENC}}} = (\textbf{ENC}_L, \textbf{ENC}_V) $ refers to the encoder structure and $\textbf{FNN}_{\theta_{\textbf{FNN}}}$ refers to the feedforward neural network. $\theta_{\textbf{class}} := \{\theta_{\textbf{ENC}},\theta_{\textbf{FNN}}\}$ denotes the parameters of our classification model. Finally, we compute the predicted event type by: 
\begin{equation}
    \widehat{e} = argmax_{e\in\mathcal{E}}P_{\theta_{class}}(e|s,v),
    \label{eq:classification}
\end{equation}

Given the support set $\mathbb{S}$, the \model architecture can be trained by minimising the cross-entropy loss:

\begin{equation}
    \mathcal{L}_{class} = \mathbb{E}_{s,v,e}[-\log P(e|s,v;\theta_{class})]
    \label{eq:class-loss}
\end{equation}

\subsection{Training in a few-shot setting}
During training, a support set of instances paired with images with K examples per event type is available, the \model model outputs the event type with the highest probability (Figure \ref{fig:pipeline}, middle).

During inference, when visual inputs are not available,
\model relies on the \im to synthesize an image given query text $x$. Then we use an image encoder to compute visual hidden presentations ${H}_{\widehat{v}}$ of the synthesized image. We then adapt the visual imagination based on the few-shot data. 

Our visual imagination module is based on Denoising Diffusion Probabilistic Models~\cite{ho2020denoising} (Figure \ref{fig:pipeline}, top). This generative model consists of a pre-trained autoencoder that maps images to a spatial latent code, a corresponding decoder that learns to map the latent representation back to the image, and a diffusion model that is conditioned on the textual input~($s$). 
Inspired by \cite{ruiz2022dreambooth}, during inference, we use the textual input of $(s,v)$ from the support set to condition the model to regenerate $v$. Furthermore, we use the input text~$s$ to synthesise visual context $\widehat{v}$. 
We then resort to the following reconstruction-based loss function to train the visual imagination module on the  synthesised image $\widehat{v}$:
\begin{align}
    \mathcal{L}_{visual} &= \mathbb{E}_{s,v,\widehat{v},\epsilon} 
    [\omega || F_{\theta_{v}}(\alpha v + \sigma\epsilon, s) - \widehat{v} ||_{2}^{2}] 
      \label{eq:visual-loss}
\end{align}
where $F_{\theta_{v}}$ denotes the conditional visual imagination. $\alpha$ and $\sigma$ are terms that control the noise schedule. $v$ is the ground truth image from the support set. $\epsilon \sim \mathcal{N}(0,I)$ is a noise term. The only trainable parameters of the visual imagination module are those of the textual encoder, and we keep the other parameters frozen.



\begin{table*}[ht]
\caption{Experimental results of few-shot multimodal event detection on M2E2 dataset. The results are reported on considering 5, 10, 15, and 20 samples of training examples.}
\begin{tabular}{lcccccccccccc}
\toprule[0.1em]
\multicolumn{1}{c}{\multirow{2}{*}{\textbf{Methods}}} & \multicolumn{3}{c}{\textbf{5-shot}}  & \multicolumn{3}{c}{\textbf{10-shot}}  & \multicolumn{3}{c}{\textbf{15-shot}}  & \multicolumn{3}{c}{\textbf{20-shot}}  \\ \cline{2-13} 
\multicolumn{1}{c}{}  & \multicolumn{1}{c}{\textbf{F1}} & \multicolumn{1}{c}{\textbf{Precision}} & \multicolumn{1}{c}{\textbf{Recall}} & \multicolumn{1}{c}{\textbf{F1}} & \multicolumn{1}{c}{\textbf{Precision}} & \multicolumn{1}{c}{\textbf{Recall}} & \multicolumn{1}{c}{\textbf{F1}} & \multicolumn{1}{c}{\textbf{Precision}} & \multicolumn{1}{c}{\textbf{Recall}} & \multicolumn{1}{c}{\textbf{F1}} & \multicolumn{1}{c}{\textbf{Precision}} & \multicolumn{1}{c}{\textbf{Recall}} \\ \hline
Text2Event~\cite{lu2021text2event} & 6.59 & 27.27 & 3.75 & 30.38 & 30.77 & 30.00 & 43.04 & 43.59 & 42.5 & 56.25 & 54.22 & 58.44 \\
Valhalla~\cite{li2022valhalla}  & 16.36  & 20.79  & 15.55  & 17.52  & 18.63  & 17.77  & 26.89  & 37.39  & 28.88  & 24.44  & 29.74  & 25.28  \\
Ferozen~\cite{tsimpoukelli2021multimodal}  & 40.36  & 43.26  & 41.11  & 42.01  & 43.21  & 42.22  & 43.19 &  44.33 &43.33& 46.11 &  47.11 & 45.97  \\
Clip~\cite{radford2021learning} & 47.93  & 50.15  & 50  & 50.38  & 55.7  & \textbf{50}  & 52.06  & 54.78  & 52.22  & 60.06  & 62.65  & 60.91 \\
\hline
\model  & \textbf{50.00}  & \textbf{50.94} & \textbf{50} & \textbf{50.58}  & \textbf{60.95} & 48.88  & \textbf{63.79}  & \textbf{70.14} & \textbf{62.22}  & \textbf{65.23}  & \textbf{67.6}  & \textbf{64.36}  \\
\toprule[0.1em]
\end{tabular}
\label{tab:main}
\end{table*}
 
In this work, we train our model using a combination of the loss functions in equations \ref{eq:class-loss} and \ref{eq:visual-loss}. 
We control the contribution of the additional visual loss by a hyperparameter $\beta$.
\begin{equation}
    \mathcal{L} = \mathcal{L}_{class} + \beta \mathcal{L}_{visual}.
    \label{eq:loss}
\end{equation}

\section{Experiments}
In this section, we describe the experimental setup used to evaluate the performance of our model, \model, and present the results of the experiments comparing \model with the baselines.

\subsection{Experimental Setup}
\textbf{Dataset.} 
We performed our experiments on the M2E2 dataset~\cite{li2020cross}, a multimodal news dataset that expands upon ACE~\cite{doddington2004automatic}. The dataset consists of 6,167 sentences, with 1,297 event mentions across 8 event types: Movement.Transport, Conflict.Attack, Conflict.Demonstrate, Justice.ArrestJail, Contact.PhoneWrite, Contact.Meet, Life.Die, and Transaction.TransferMoney. To evaluate the performance of our model with limited training data, we considered a few-shot setting where only \emph{K} (5, 10, 15, and 20) data points with both text and image pairs were used during the training session. Furthermore, during inference, only text was provided to the model, and paired images were discarded.

\textbf{Evaluation Metrics.} Following previous works in addressing this task~\cite{li2020cross}, the results are reported using macro-averaged Precision (P), Recall (R) and F-measure (F-1) score metrics. 

\textbf{Baselines.} In order to validate the proposed  model through experimental comparison, we selected the following multimodal event detection models as the baselines:
\begin{itemize}
\item Text2Event~\cite{lu2021text2event}: A text-only baseline that trains a pre-train language model (T5) without any visual information. 
\item Valhalla~\cite{li2022valhalla}: A multimodal sequence-to-sequence model for machine translation by leveraging visual hallucination at test time. We train Valhalla by training the model to generate only the event type.
\item Frozen~\cite{tsimpoukelli2021multimodal}: A method for transforming large language models into multimodal few-shot
learning systems by extending the soft-prompting philosophy of prefix tuning.
\item Clip \cite{radford2021learning}: A recent
model, based on Contrastive Language-Image Pre-training, learns a multi-modal embedding space, which can be used to estimate the semantic similarity between a given text and an image.  
\end{itemize}

\textbf{Experimental Settings.}  
We initialized our Visual Imaginator using the parameters of the pre-trained stable diffusion model~\cite{Rombach_2022_CVPR}. For text and image encoding, we employed CLIP ViT-L/14~\cite{radford2021learning} with an embedding size of 768 and a dropout of 0.3 over the textual encoder. We use the Huggingface transformer library~\footnote{\url{https://github.com/huggingface/transformers}} to implement the \im and the encoder of our model. Optimization was performed using Adam~\cite{kingma2014adam} with a learning rate of $2e-5$. We used a 1-layer feedforward neural network, and set $\beta$ to 0.01. Our model was trained for 50 epochs with a batch size of 4. For the baselines, we used the source code and implementation details provided by the authors~\footnote{As we were unable to access the source code of the Frozen model, we implemented the model by closely following the descriptions provided in the paper. This ensured that our implementation was as close as possible to the original model.}. We prepared the data for all models in the same way.


\begin{figure*}[ht]
    \centering
    \resizebox{.90\textwidth}{!}{
    \includegraphics{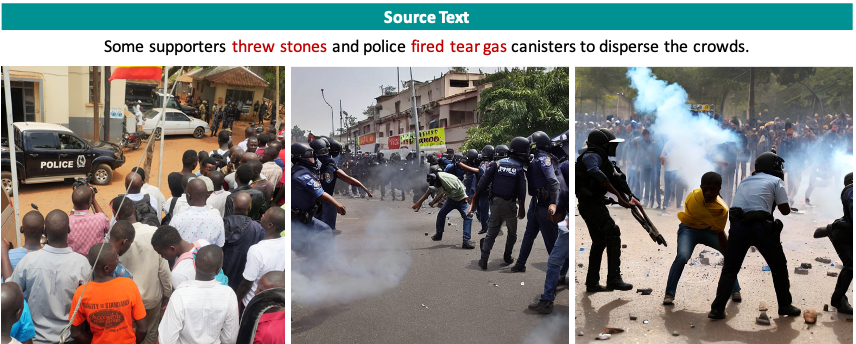}}\\
    \vspace{0.2 cm}
    \resizebox{.90\textwidth}{!}{
    \includegraphics{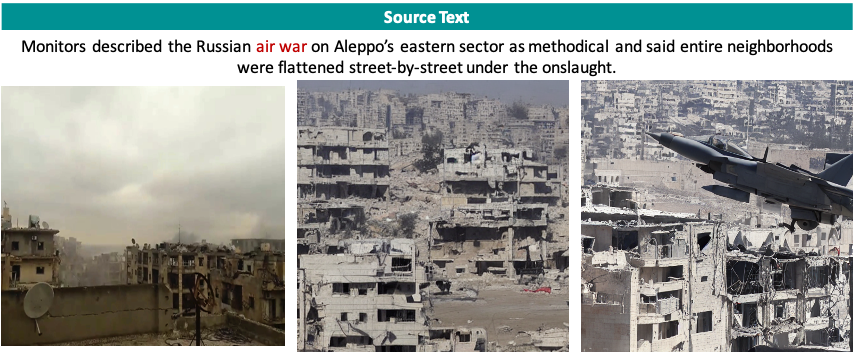}}
    \caption{A qualitative comparison of the ground-truth images to synthesised images. \textbf{Left:} the original image. \textbf{Middle:} the image generated by the pre-trained stable diffusion model~\cite{ho2020denoising}. \textbf{Right:} the image synthesized by our customized visual imaginator module \model.}
\label{fig:imagination}
\end{figure*}

\subsection{Observations and Discussions}
Table~\ref{tab:main} shows the performance of the models on the M2E2 dataset, where at the training time only \emph{K} data points with both modalities are provided. The superiority of our model can be observed from the F1, precision and recall scores associated with event type detection. On F1 score, our model achieves a substantial 43\% improvement on 5-shot and a 9\% improvement on 20-shot over the text-only model Text2Event~\cite{lu2021text2event} on event type detection. Our \model model significantly outperforms the text-only baselines on all four few-shot settings, which demonstrates the effectiveness of visual imagination for text-only. This result also indicates that when the number of training datasets are less (i.e., 5), unimodal, text-only models fail to extract information from events and fusing visual information during inference, results in poor detection of event types. 
Our proposed approach, as shown in Table~\ref{tab:main}, improves on all compared methods, achieving best precision, recall and F1 scores under different few-shot settings in all but one metric (Recall in 10-shot).

In summary, these consistent improvements clearly show that visual imagination can effectively leverage visual representation from the source sentences at test time for improved event detection.

\subsection{Qualitative Analysis of the Visual Imaginator}

In this section, we present two samples of images generated by our model. We randomly selected two data points from the test set of M2E2 and used the textual information to condition our visual imagination to generate the visual representation of the text (Figure 4). As can be seen in these examples, while the ground-truth images fail to fully portray the textual information, our proposed model incorporates this information in the visual representation. For instance, in the first example, the original image fails to provide any visual clues about \emph{threw stone} or \emph{fire/smoke}. In contrast, our proposed \im captures this information (Figure 4, first row). Similarly, in the second example, \emph{air war} is properly portrayed in the image generated by our model, whereas it is missed in the ground truth image as well as the image generated by stable diffusion (Figure 4, second row). It can be observed that sometimes the synthesized images are more informative compared to their original counterparts. This indicates the ability of our \model model to generate context-relevant images during inference.


\begin{table*}[ht]
\caption{Ablation study on M2E2 dataset: Effects of the visual and textual modality and the \im on the performance of our model in few-shot event detection. \model~(\texttt{L}): no images. \model~(\texttt{V}): no text. \model~(\texttt{RET}): image-retrieval.}
\begin{tabular}{lcccccccccccc}
\hline
\multicolumn{1}{c}{\multirow{2}{*}{\textbf{Methods}}} & \multicolumn{3}{c}{\textbf{5-shot}}  & \multicolumn{3}{c}{\textbf{10-shot}}  & \multicolumn{3}{c}{\textbf{15-shot}}  & \multicolumn{3}{c}{\textbf{20-shot}}  \\ \cline{2-13} 
\multicolumn{1}{c}{}  & \multicolumn{1}{c}{\textbf{F1}} & \multicolumn{1}{c}{\textbf{Precision}} & \multicolumn{1}{c}{\textbf{Recall}} & \multicolumn{1}{c}{\textbf{F1}} & \multicolumn{1}{c}{\textbf{Precision}} & \multicolumn{1}{c}{\textbf{Recall}} & \multicolumn{1}{c}{\textbf{F1}} & \multicolumn{1}{c}{\textbf{Precision}} & \multicolumn{1}{c}{\textbf{Recall}} & \multicolumn{1}{c}{\textbf{F1}} & \multicolumn{1}{c}{\textbf{Precision}} & \multicolumn{1}{c}{\textbf{Recall}} \\ \hline
\model~(\texttt{L})  & 46.19 & 49.83 & 45.55 & 50.26 & 60.86 & 47.77 & \textbf{64.09} & 70.02 & 62.13 & 62.65 & 64.85 & 63.21 \\
\model~(\texttt{V})  &  17.40 & 16.88 & 20.0 & 16.57 & 22.39 & 18.88 & 14.70 & 18.23 & 14.44 & 17.23 & 29.07 & 17.24 \\
\model~(\texttt{RET})  & 39.32 & 40.97 & 41.11 & 50.17 & 57.67 & 50.0 & 51.0 & 59.44 & 48.88 & 54.76 & 54.44 & 56.32 \\
\hline
\model & \textbf{50.00}  & \textbf{50.94} & \textbf{50} & \textbf{50.58}  & \textbf{60.95} & \textbf{48.88}  & 63.79  & \textbf{70.14} & \textbf{62.22}  & \textbf{65.23}  & \textbf{67.6}  & \textbf{64.36} \\ 
  \hline
\end{tabular}
\label{tab:ablation}
\end{table*}

\subsection{Ablation study}
We conducted an ablation study to investigate the impact of each modality in few-shot event detection. We evaluated the performance of our model under two ablation settings: (1) when the visual context was fully disregarded during training~(\model~$(\texttt{L})$), and (2) when the textual context was disregarded during training~(\model~$(\texttt{V})$). Moreover, we replaced our proposed Visual Imaginator with an image retrieval method~\cite{wu2021good} and evaluated the performance of our model under this setting~(\model~$(\texttt{RET})$).

Table \ref{tab:ablation} summarises the results of our ablation study. As shown, \model~$(\texttt{L})$ outperforms \model~$(\texttt{V})$ in all settings, indicating that textual modality carries richer information than visual modality for event detection. However, using both modalities outperforms models trained on only one modality (i.e.\ text), across all metrics for different support sets~(with the exception of the F1-score for the 15-shot setting). This observation demonstrates the complementary effect of visual context for event detection. Moreover, \model~$(\texttt{RET})$ performs worse than, but close to \model~$(\texttt{L})$, demonstrating the effectiveness of our model when an image retrieval model is used instead of the \im. Finally, our full model with \im outperforms \model~$(\texttt{RET})$ in all settings, with a margin of up to 12 points in F1-score. This observation emphasizes the importance of our proposed \im in synthesizing context-related information.

\section{Conclusion }
This work addresses the challenges of event detection when the model faces low-resource multimodal data, during training and inference. We propose a few-shot visually-fused model which combines a pre-trained multimodal module with a customised \im module. Using the few-shot learning approach and leveraging only a few labelled paired image-text samples, our model adapts \im  to a scenario of interest, thereby reducing the domain shift between the original domain of the pre-trained models and the target domain. The customised \im module also allows for effective event detection when the visual modality is absent by providing relevant visual context to the model during inference. Our experiments demonstrate that our proposed approach effectively exploits the low-resource training data, and outperforms state-of-the-art techniques for event detection.

\bibliographystyle{IEEEtran}

\bibliography{IEEEabrv}

\end{document}